\documentclass[letterpaper, 10 pt, conference]{ieeeconf}  

\IEEEoverridecommandlockouts                              

\usepackage{graphicx}
\usepackage{graphics} 
\usepackage{tabularx}
\usepackage{float}
\usepackage{color}
\graphicspath{{./figure/}}

\title{\LARGE \bf
Origami-inspired Bi-directional Actuator with Orthogonal Actuation
}

\author{Shuai Liu*, Sheeraz Athar*, and Michael Yu Wang, \IEEEmembership{Fellow, IEEE}%
\thanks{*The authors have equally contributed to this work.}%
\thanks{All authors are with the Department of Mechanical and Aerospace Engineering, The Hong Kong University of Science and Technology. Emails: {\tt\small sliubw@connect.ust.hk, sathar@connect.ust.hk, mywang@ust.hk}.}%
}

\begin{document}
\maketitle
\thispagestyle{empty}
\pagestyle{empty}

\begin{abstract}
Origami offers a promising alternative for designing innovative soft robotic actuators. While features of origami, such as bi-directional motion and structural anisotropy, haven't been extensively explored in the past, this letter presents a novel design inspired by origami tubes for a bi-directional actuator. This actuator is capable of moving in two orthogonal directions and has separate channels throughout its body to control each movement. We introduce a bottom-up design methodology that can also be adapted for other complex movements. The actuator was manufactured using popular 3D printing techniques. To enhance its durability, we experimented with different 3D printing technologies and materials. The actuator's strength was further improved using silicon spin coating, and we compared the performance of coated, uncoated, and silicon-only specimens. The material model was empirically derived by testing specimens on a universal testing machine (UTM). Lastly, we suggest potential applications for these actuators, such as in quadruped robots.

\end{abstract}

\section{INTRODUCTION}
Origami, the ancient Japanese art of paper folding, has become a new norm in engineering design \cite{rus2018design}. The ease of manufacturing, compliance, and compactness are a few of the Origami features that have attracted the scientific community's attention. Scientists and Engineers are now using Origami techniques in various applications ranging from surgical instruments \cite{boyvat2017addressable} to space exploration \cite{wilson2013origami}. Recently, origami-inspired robots \cite{yao2019reconfiguration}; actuators \cite{peraza2014origami} and other miscellaneous items are also proposed \cite{defigueiredo2019origami, lee2013deformable}.

One of the main areas of origami application in robotics is soft robotic actuators. Through fast and effective bending, along with the hinge, origami actuators can produce complex motions. Many robotics design, utilizing the collapsing and expanding behavior of origami, have been developed in the past including crawling \cite{koh2012omega}, flying \cite{sreetharan2012monolithic}, gripping \cite{firouzeh2017under}, locomotion \cite{weston2017towards} etc. One study aimed at developing forceps-like origami gripper for use in surgical equipment \cite{edmondson2013oriceps}. These grippers work well in small-scale applications only, and higher loads result in slagging of the forceps. \cite{santoso2020origami} Proposed a design for tendon routed continuum origami robotic arm. Tendon-driven manipulators are effective due to their agility and compliance. However, due to multiple driving strings, their control is often cumbersome. In addition to these shortcomings, most of the studies underutilized the origami feature of movement in multiple directions.

 \begin{figure}[thbp]
      \centering
      \includegraphics[width=\columnwidth]{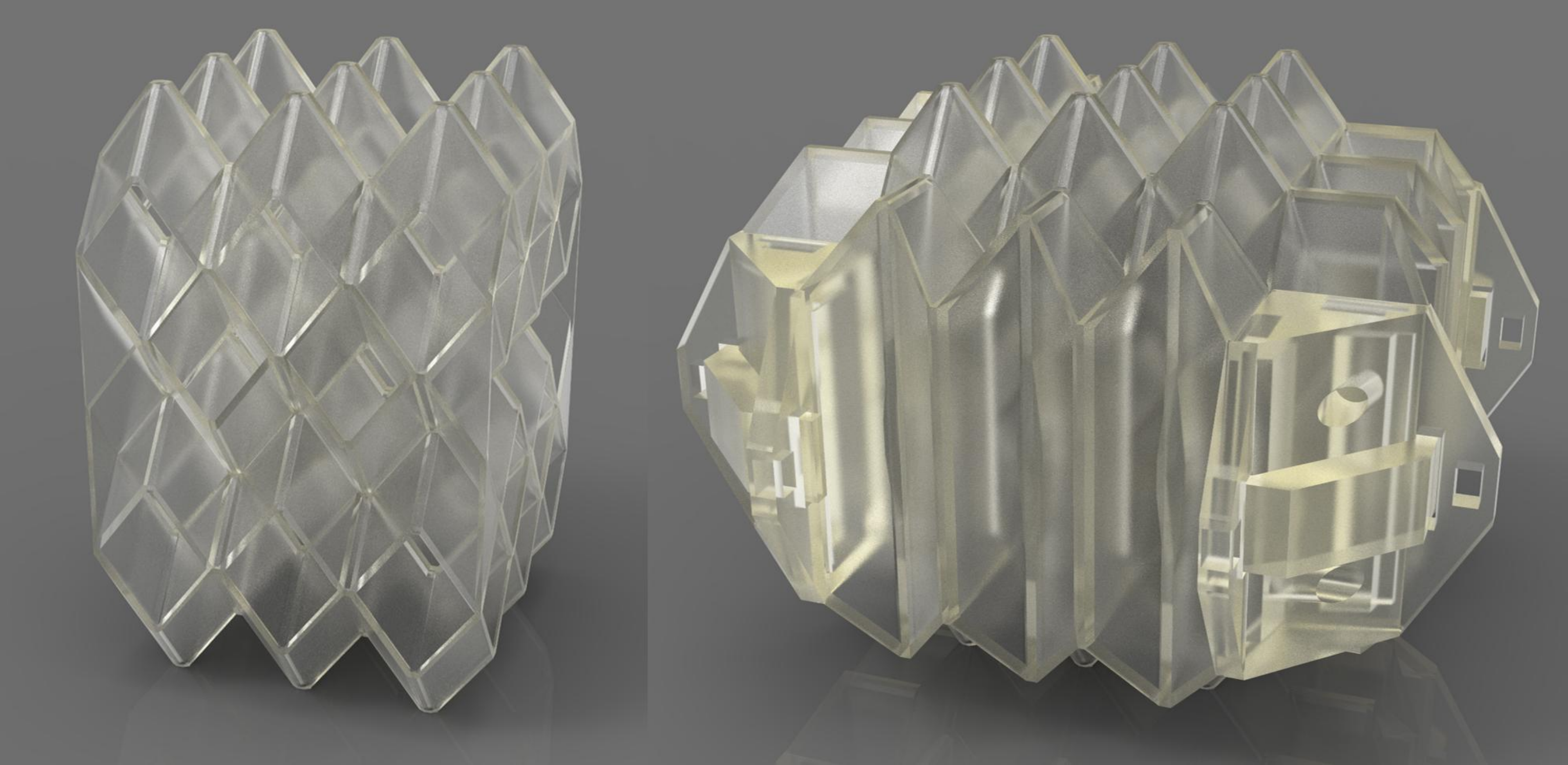}
      \caption{Rendering of the Bi-directional pneumatic actuator}
      \label{Rendering}
\end{figure}

Bi directionality is one of the significant advantages of Origami design. Through successive folding and unfolding, Origami structures can morph into different shapes. This benefit of the Origami was utilized in the deployment of the solar panels in space \cite{10.1115/DETC2013-13490}. \cite{kim2018origami} used the same pattern with elastomeric material and developed a dual-morphing origami structure. \cite{li2017fluid} Employed origami folding into fluid-driven artificial muscles. These works have demonstrated the potential benefit of implementing transforming origami structures into robotic components. However, the development of this technique into multi-directional actuators is not widely proposed. 

To adopt this unique feature in real-world applications, we here in this paper are proposing a design for a bi-directional origami actuator inspired by origami tubes \cite{filipov2015origami}. We have also developed a robot from this actuator to showcase its utility. Traditionally origami tubes are only used in the literature for passive applications or as a structural component. \cite{filipov2016origami} Has reported the design and structural analysis for the origami tubes with different cross-sections. Analysis of an extended origami tubes family is also presented in \cite{chen2017extended}.
Nevertheless, neither of these talked about adapting an origami tube in an actuator that can be actively controlled. Our study presents such an implementation, which will be very helpful in developing novel robotic applications. In our design, we have arranged the origami tubes in a 3D arrangement, which can deform in two directions while resisting the motion in the third one. As a result, our actuator possesses a unique variable stiffness or anisotropic property, which can help adjust the design according to the task at hand. 

The remaining paper is arranged in the following manner. Section II describes the design methodology of the proposed actuator and fabrication process. Modeling of the actuator, materials and the characterization of the actuator is presented in section III. The possible applications are discussed in section V. Section VI contains a discussion on the actuator's performance and results.

\section{DESIGN OF THE ACTUATOR }

\subsection{Design}
\begin{figure}[thbp]
      \centering
      \includegraphics[width=\linewidth]{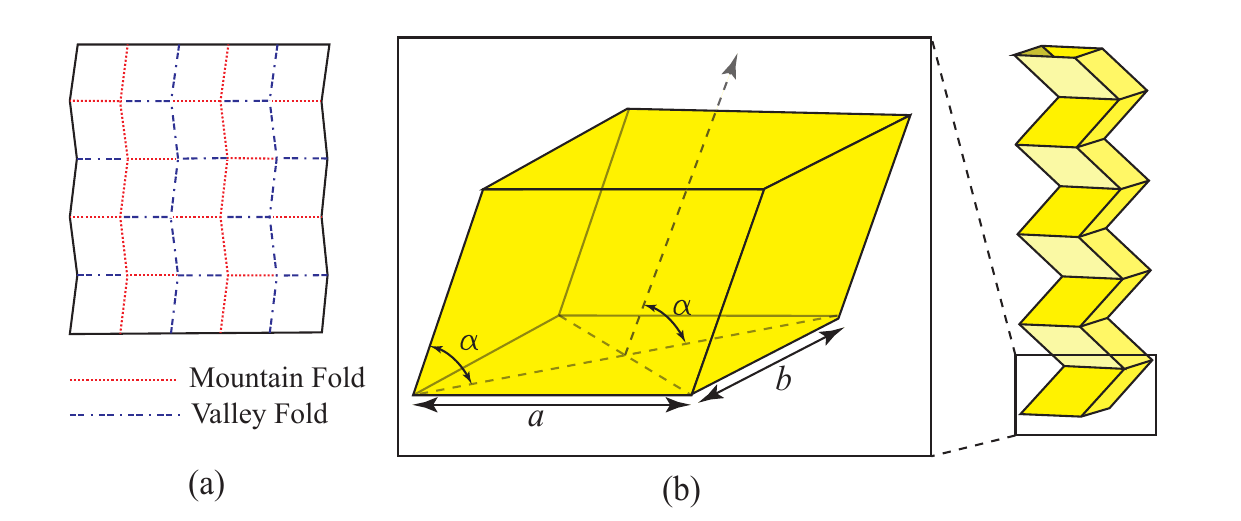}
      \caption{(a) Miura-Ori crease pattern, (b) Design parameters of our single origami tube inspired from Miura-Ori pattern}
      \label{Miura}
\end{figure}

\begin{figure}[thbp]
      \centering
      \includegraphics[width=\linewidth]{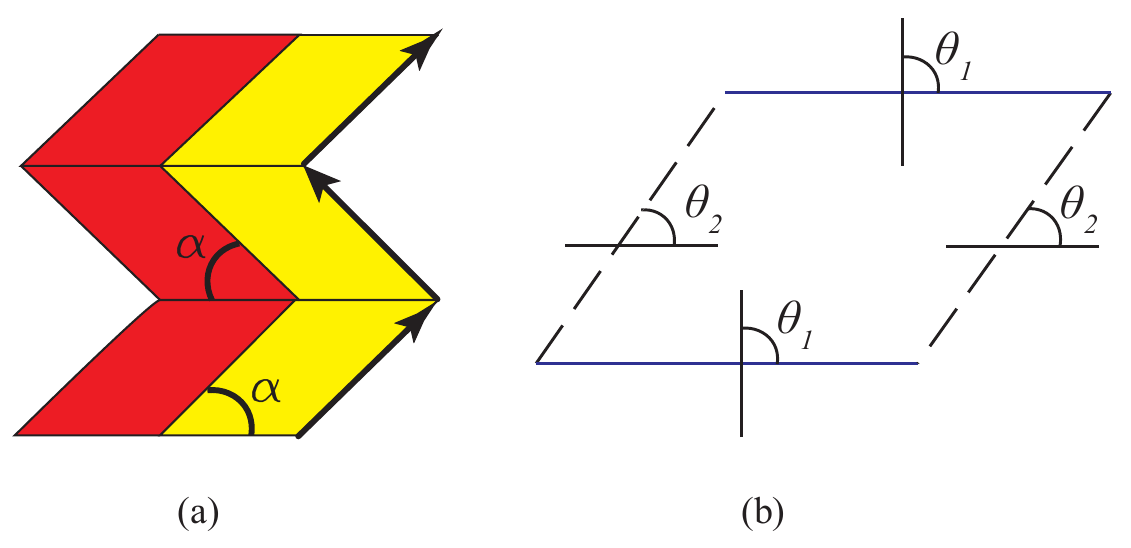}
        \caption{(a) Projection used for the origami tube design, (b) Cross-section of the origami tube used in the study}
        \label{cross-section}
\end{figure}

We will firstly discuss the design principle of primary tubes and how their parameters influence the performance. Cross-section and projected shape play a vital role in the movement of the origami tubes. According to the translational symmetry methods, \cite{filipov2016origami} admissible cross-section for an origami tube must have a presence of corresponding edge groups with the same slope and length on either of its portion. In other words, there should be groups of parallel sides in the cross-section, and their total length should also be the same. 

Numerous cross-sections are possible to construct that satisfy these conditions. However, we have only discussed the quadrilateral cross-section tubes in the study due to their simplicity, effectiveness, and ease of manufacturing. Besides, different cross-section analysis is not essential to my research. The main objective here is to develop actuators that can be used for real-world applications. Varying the cross-section does not provide any added advantage on this front, and no matter how much we change the shape, possible motion directions and movements remain the same. Therefore my thesis focuses on quadrilateral cross-sections only; hitherto, the concept discussed can be extended to all the possible cross-sections. 

Fig.\ref{cross-section} shows the quadrilateral cross-section geometry, with $\theta_1,\theta_2$ being the slope angles of the parallel edge groups. The choice of these angles is arbitrary, and just as a use case, we have selected the values for a perfect square. To convert the cross-section into a 3D figure, we need to make projections. There are several ways of projecting 2D shapes to make a tube, but not all of them produce the proper foldability required for the movement. The projection process used in the current study is shown in Fig.\ref{cross-section}(a), which yields linear motion. Other projections yielding rotary and spiral motion are also possible to construct. Here we have only discussed the design process for linear motion so that discussion doesn't become overwhelming. However, the principles presented are equally applicable to all projection classes.

Design parameters of a single tube structure are depicted in Fig.\ref{cross-section}. It can be seen that the Miura-Ori origami pattern inspires the folding pattern. The origami tube has a repetitive design; therefore, defining the parameters on one unit will be sufficient. $a$ and $b$ are the dimensions of tube cross-section; as mentioned earlier, we have modeled the cross-section as a square for simplicity, hence the value of $a$ and $b$ is identical throughout the study. $\alpha$ is the angle of projection. It governs the motion of the cross-section along the projected path and plays a crucial role in the movement of the tubes. If the angle $\alpha$ is significantly less, the motion will be stalled due to weakened action, and with a very high value, the tubes will not be connected well. We have set the value of $\alpha$ as 45$^\circ$ for the sake of symmetry. For the other requirements, these parameters can be architectured accordingly.

\begin{figure}[thpb]
      \centering
      \includegraphics[width=\linewidth]{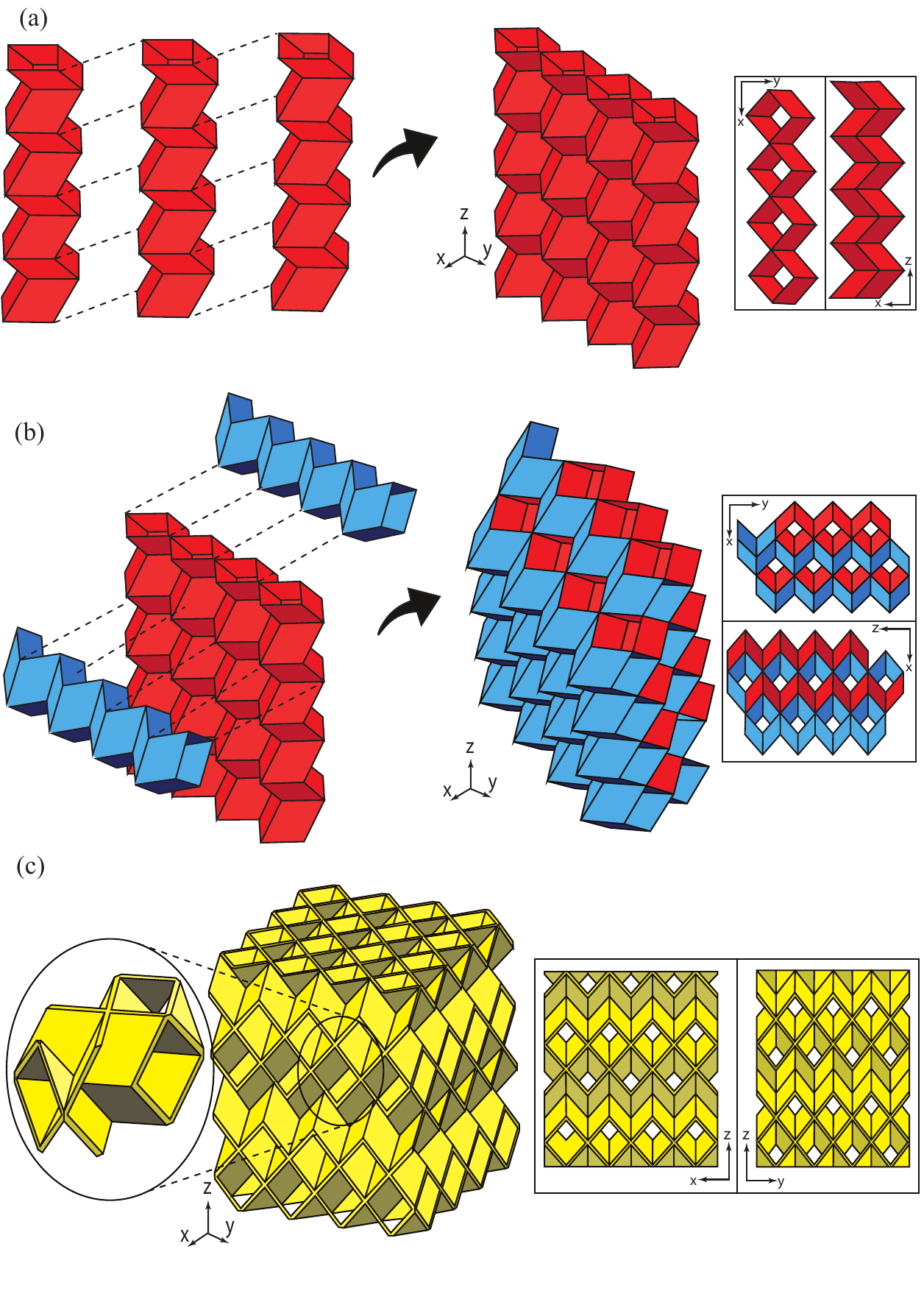}
      \caption{Stages of the actuator design: (a) Tubes in vertical orientation are joined together end to end, (b) Tubes in horizontal direction are then connected, (c) Final design constructed from the unit cell shwon in circular inset}
      \label{Design}
\end{figure}

The final step in the design process is the assemblage. Fig.\ref{Design} depicts details of different stages in the assemblages process. In the first stage, origami tubes in vertical orientation are connected edge to edge. These connected tubes are then sandwiched between origami tubes in the horizontal configuration. This process is repeated till we get enough tube structures required for the actuation. In the last stage, the unit structure is extracted from the middle of the design and then patterned on all sides to generate a symmetric tubing structure. After this step, the design we get is not closed from all sides and requires further post-processing to make it airtight and suitable for robotic applications.

\begin{figure}[t]
    \centering
    \includegraphics[width=\linewidth]{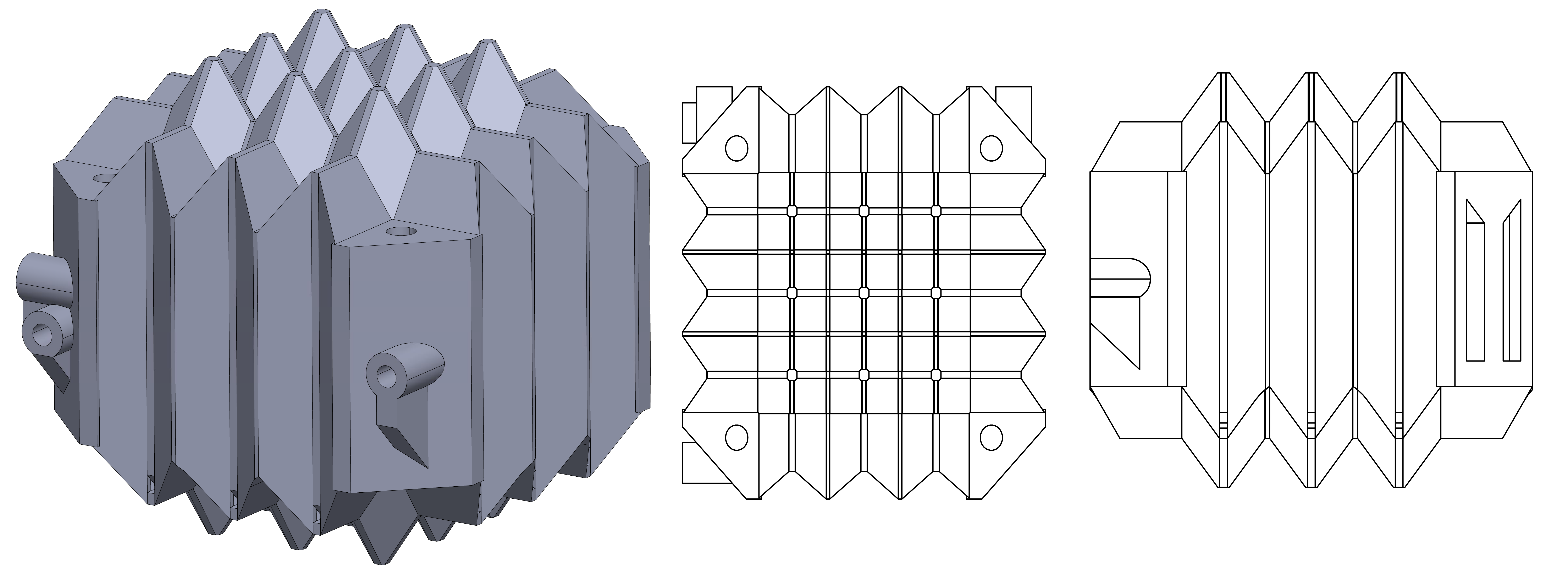}
    \caption{CAD model of the bi-directional actuator along with top view and side view}
    \label{CAD_bidir}
\end{figure}

\begin{figure}[t]
    \centering
    \includegraphics[width=\linewidth]{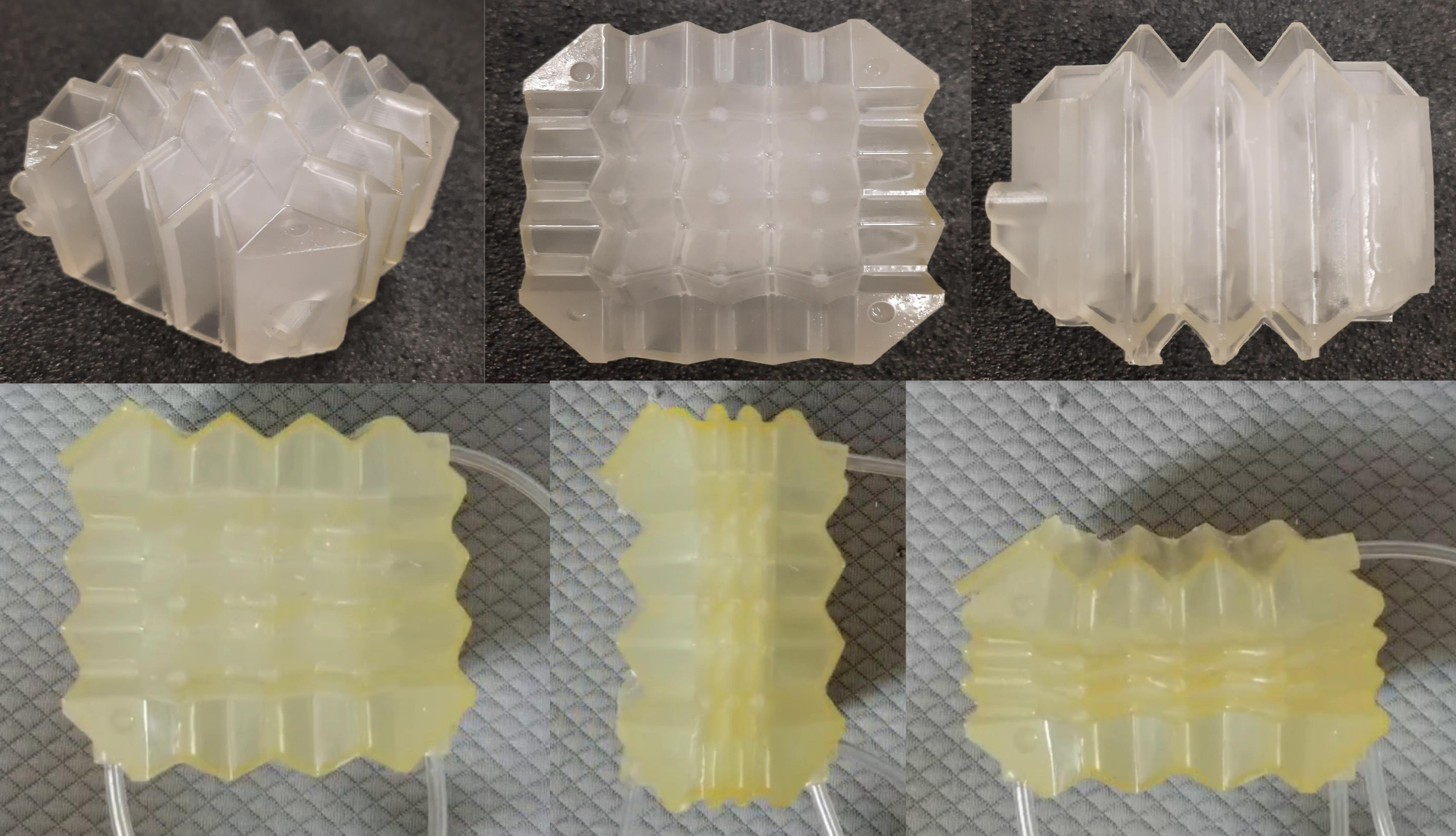}
    \caption{Photos in the first row shows different views of the real actuator fabricated using LCD method. Second row illustrates motion capacity of the actuator in two orthogonal directions}
    \label{fab_pics}
\end{figure}

To make the structure airtight for pneumatic actuation, we attached auxiliary elements shown in Fig\ref{CAD_bidir}. While providing the necessary fit for pneumatic supply, these structure doesn't interfere with the motion of the actuator. To attach different fixtures to the design, we further changed the shape and made several holes to connect the fixture components. In the end, the design is capable of showcasing linear motion in two mutually perpendicular directions (Fig.\ref{fab_pics}). The structure also restricts movement in the third direction and hence possesses anisotropic properties and can be used as a variable stiffness component. The motion generated is also unique and can result in various ingenious applications, including crawling robots, variable-length gripper, etc.

The printed version of the actuator is shown in Fig\ref{fab_pics}. We can notice the high-quality surface finish resulting from LCD and SLA printing methods. The resin material is transparent after print, and therefore it is easy to visualize inner layers. The working of the printer is illustrated in the second row of Fig.\ref{fab_pics}. The first two images show linear motion in one direction, and the third image depicts translation in the other orthogonal direction. Photographs are taken when the actuator is vacuumed at 98KPa using a small portable pump. These images prove the motion concept and capability of the actuator.

\begin{figure*}[thpb]
      \centering
      \includegraphics[width=\linewidth]{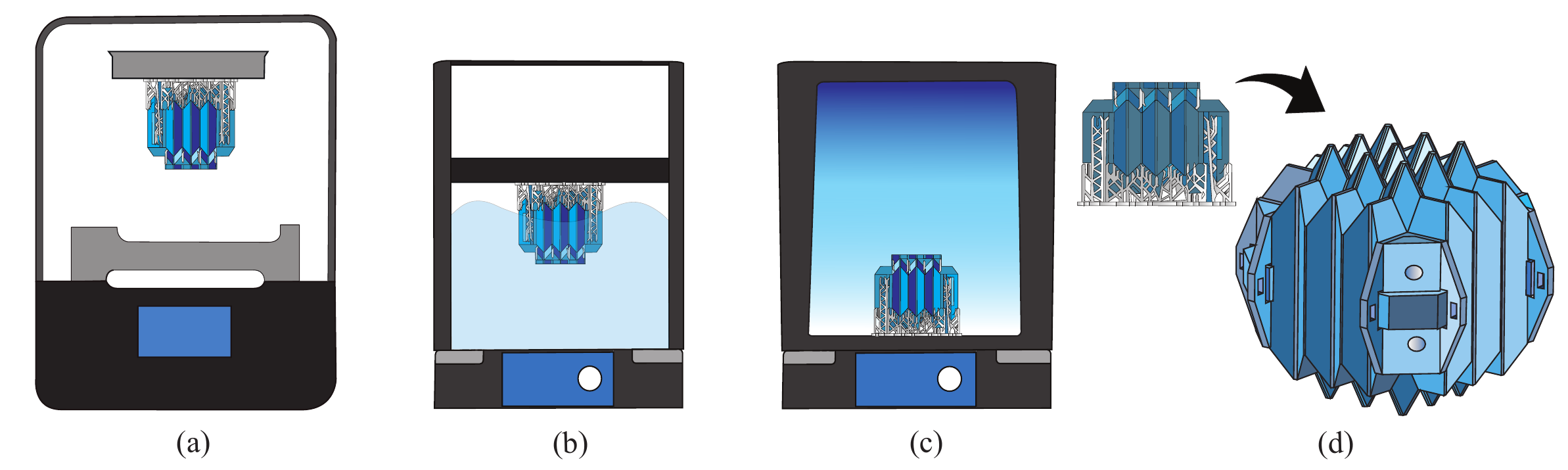}
      \caption{Fabrication Process: (a) SLA printing, (b) Alcohol Wash, (c) Curing, (d) Actuator after curing and removal of supports, (e) Coating of silicon on the actuator, (f) Heating in the oven to cure the silicon, (g) Final Actuator}
      \label{fabrication}
\end{figure*}

\subsection{Fabrication}

Fabricating the proposed origami actuators using rapid prototyping methods is one of the main contributions of the current study. Here, we have discussed two fabrication approaches: Stereolithography (SLA) and Liquid Crystal Display (LCD). Both are commonly used 3D printing techniques for printing with resin materials and provide their own merits and demerits. This section will briefly mention the main benefits of these methods in fabricating the proposed design. We will also discuss possible ways to circumvent commonly faced problems.  

\begin{table*}[t]
 \centering
    \caption{Properties of Different Material tested for fabrication}
    \label{material1}
\begin{tabularx}{\linewidth}{|X|X|X|X|}
\hline
\textbf{Material} & \textbf{Shore Hardness} & \textbf{Tear Strength} & \textbf{Tensile Strength}   \\ \hline
Resinone F39 T    & 60-75A                  & 47.2 kN/m              & 7.9 MPa                      \\ \hline
Formlabs Elastic  & 40-50A                  & 19.1 kN/m              & 3.23 MPa                     \\ \hline
Formlabs Flexible & 80-85A                  & 13.3-14.1 kN/m         & 7.7-8.5 MPa                  \\ \hline
Hesu H04          & 35-45A                  &     NA                   & NA                             \\ \hline
Hesu H03          & 50-60A                  &     NA                 &   NA                            \\ \hline
Resinone F80      & 58A                     & 5.7 kN/m               & 1.7 MPa                      \\ \hline
Carbon SIL30      & 35A                     & 10 kN/m                &   NA                           \\ \hline
LUVOSINT X92A-2   & 88A                     &    NA                   & 20-15 MPa                    \\ \hline
\end{tabularx}
\end{table*}

\begin{table*}[h]
 \centering
    \caption{Properties and Problems faced with different materials}
    \label{material2}
\begin{tabularx}{\linewidth}{|X|X|X|X|}
\hline
\textbf{Material} & \textbf{Elongation at break} & \textbf{Viscosity} & \textbf{Problems}     \\ \hline
Resinone F39 T    & 255.1\%                       & 980mpa.s           & Hard to post-cure     \\ \hline
Formlabs Elastic  & 160\%                         &   NA                 & Brittle and too thick \\ \hline
Formlabs Flexible & 75-85\%                       &   NA                 & Brittle               \\ \hline
Hesu H04          &  NA                            & 320-550mpa.s       & Not durable           \\ \hline
Hesu H03          &  NA                            & 280-450mpa.s       & Not durable           \\ \hline
Resinone F80      & 80\%                          & 2300 mpa.s         & Elongation too small  \\ \hline
Carbon SIL30      & 350\%                          &    NA                & Very Expensive           \\ \hline
LUVOSINT X92A-2   & 500\%                         &    NA                & Powder                \\ \hline
\end{tabularx}
\end{table*}

\begin{figure}[t]
\centering
\includegraphics[width=\linewidth]{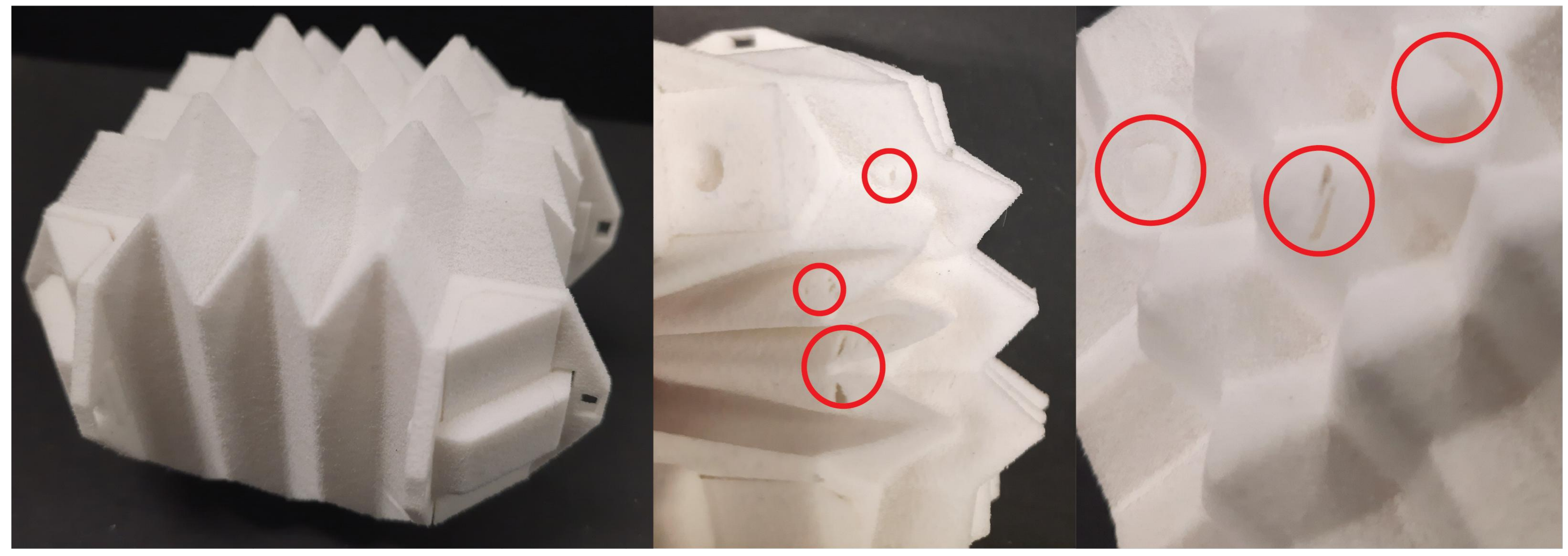}
\caption{SLS printed parts. Two images on the left show defects circled in red }
\label{sls}
\end{figure}

Both SLA and LCD are resin 3D printing methods and fabricate parts by exposing the resin to a light source. In SLA, the light source is a laser, and in LCD, an LCD screen is used to cure the resin material. Therefore, both methods are very similar in principle, with the only difference being the light source. Most of the SLA desktop 3D printers contain a resin tank fitted with a transparent surface at the base. This layer is non-stick and acts as a substrate for the liquid resin to cure against, thereby allowing easy detachment of successively printed layers. Right above the tank is a build platform that descends into the resin tank for every printing layer. Once it dipped, a laser points at two mirror galvanometers, directing the light to the desired coordinates. The laser then follows all the layers in the STL file to print the whole structure.  

Like SLA printing, LCD also has parts like a resin tank, bottom transparent surface, and a build platform. As mentioned above, the main difference between the two is a light source. While SLA uses lasers, LCD employs an LCD screen. Using the screen instead of the laser allows the LCD printers to cure the whole layer in one go, and therefore they are faster than SLA when printing large parts. However, for small parts, the SLA method will be more time-efficient. LCD printer prints in terms of 3D pixels known as voxels that also govern the printing process's resolution.
Contrary to this, SLA printers have round laser spot which specifies their resolution. Both methods are way better than other 3D printing methods, such as FDM, in terms of accuracy and precision. Precision is quantified by the uniformity of the different layers printed. In SLA, there is only one laser that is curing the resin; therefore, uniformity is relatively high. In LCD, the screens have issues of non-uniform projection of the light, which often requires calibration. The surface finish of the parts manufactured using both methods is also comparable and best in class. 

We used both methods for fabricating the actuators. Form3 printer from Formlabs, USA is used to print parts using SLA technique, and JX215 3D printer is used for LCD method. One of the crucial elements of resin 3D printing is the selection of proper resin material. In our study, we have discussed nine resin materials from different suppliers with problems faced in each. Various properties of the materials used and their associated issues are listed in tables \ref{material1} and \ref{material2} (NA in the table means data is not available from the supplier). Among these, two materials from formlabs are used in SLA while the rest are used in LCD. For the proper working of the actuator, a material that is flexible enough to produce the motion and strong enough to resist rupture in the inside structure is required. Parts printed with SLA are flexible, but they were not durable and broke after few actuation cycles. The same was the case with Hsu materials on LCD printers; they were also flexible but not durable. Finally, we tested Resinone F39T material which had properties of both flexibility and durability. The only problem with this material was post-curing, which we tackled using a high-power UV light source. 

The whole fabrication process for SLA and LCD methods are described in Fig.\ref{fabrication}, respectively. SLA printing typically takes three steps to complete, namely printing, washing, and curing. The printing step is done using SLA printer Form3 by Formlabs, USA (Fig.\ref{fabrication}(a)). After the printing, there remains un-cured resin inside the holes, which needs to be washed out before proceeding further. For this, printed parts are dipped in an ethyl alcohol bath (Fig.\ref{fabrication}(b)). Ethyl alcohol scrubs away excess resin and cleanses the surface. Following this, printed parts are further cured under light radiation to maximize the material properties (Fig.\ref{fabrication}(c)). This method, known as post-curing, is done using a Formlabs machine called Form Cure. Once the part is post-cured, supports are removed, and different regions are attached (Fig.\ref{fabrication}(d)). Similar to SLA, the LCD method also has steps of printing, washing, and post-curing. In LCD, washing was done by dipping in Isopropyl Alcohol (IPA), and curing was done using a UV light source. We used UV light because Resinone F39T material was difficult to post-cure and required a high energy supply for the curing. 

In addition to these methods, we also used the selective laser sintering (SLS) printing method for fabricating the actuator. In SLS, successive printing layers are printed by selectively sintering the powdered material. Due to the presence of extra material on all sides of the print, the SLS method requires virtually no support structures for printing. We used powdered TPU material for the fabrication; results are shown in Fig.\ref{sls} with defects circled in red. The main drawback of this method is poor structural strength caused by weak layer adhesion and micro holes. The resulting structure is fragile, and breakage occurs even after few actuation cycles. Different coating processes can mitigate this, but we left this part in the discussion due to the involved complexity.

\section{Modelling and Characterization}
\subsection{Characterization}
\subsubsection{Force Characterization}
\begin{figure}[t]
    \centering
    \includegraphics[width=\linewidth]{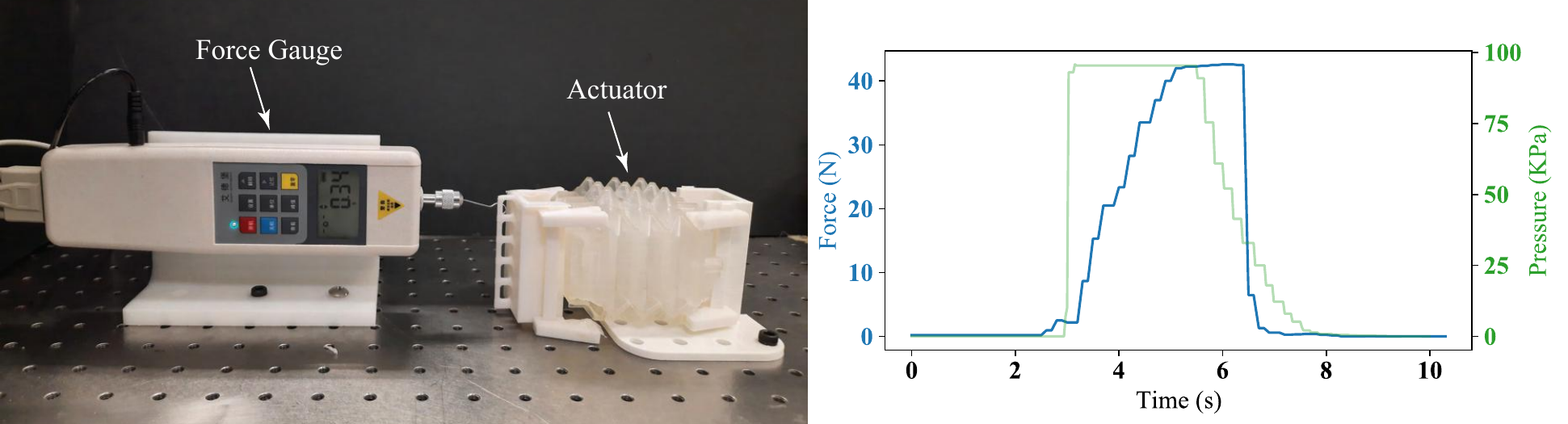}
    \caption{Force characterization of the bi-directional actuator: on the left is the setup used to record data and on the right is the result of the experiment}
    \label{force_bidir}
\end{figure}

For better quantification of actuator working, knowledge of its force profile is imperative. Therefore, we did experiments to record the maximum amount of force exerted by the actuator in both actuation directions. The force is measured using a digital force gauge attached to the end of the actuator using fixtures depicted in Fig.\ref{force_bidir}. To minimize the error in recording, we have placed the force gauge in close proximity of the actuator along the actuation line. The result of the force characterization is shown in the plot of Fig.\ref{force_bidir}. It can be seen that the maximum force exerted by the actuator is around 42N when vacuum by -94kPa. Also, we can see that pressure increase is instantaneous which means application of pressure doesn't impact the force generated by the actuator. 

\begin{figure}[t]
    \centering
    \includegraphics[width=\linewidth]{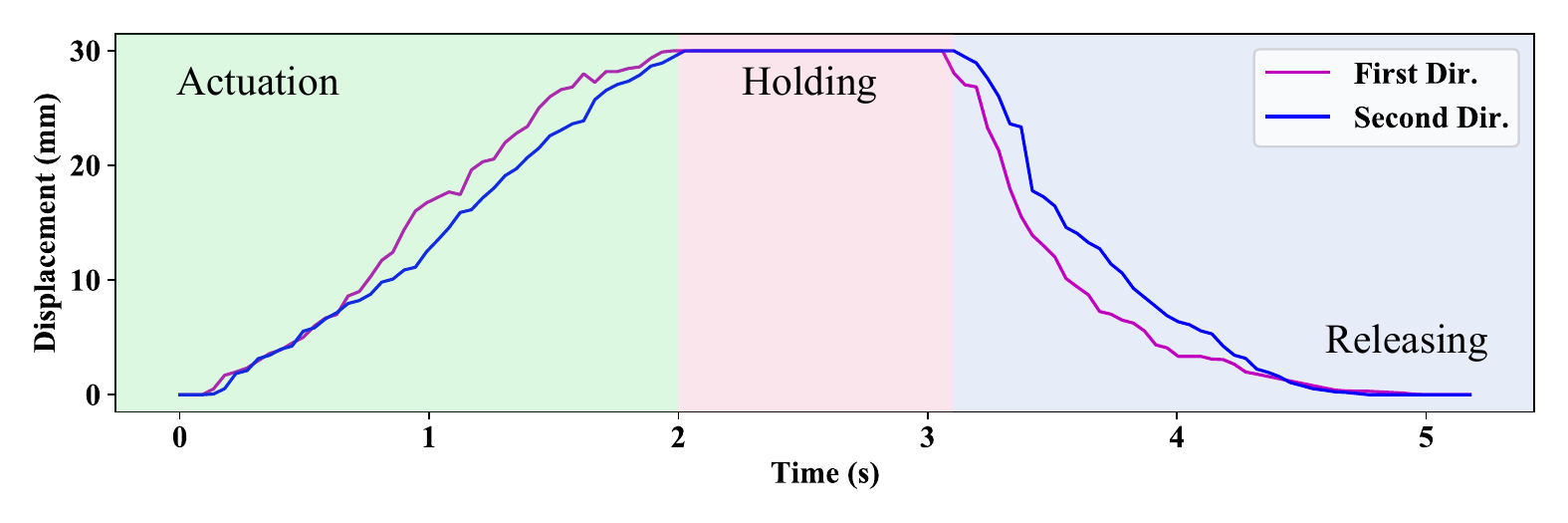}
    \caption{Step Response of the bi-directional actuator in both directions}
    \label{step_bidir}
\end{figure}

\subsubsection{Step Response}

Step response of the actuator is analyzed on the action of one actuation cycle. The step response is essential in predicting the behavior of the actuator and acts as a special design element. Bi-directional actuators printed using LCD methods are tested in these experiments by applying constant vacuum pressure of -50KPa. Visual data is recorded and then analyzed using video analysis software for the extraction of desired parameters. The step response graph of the actuator is shown in Fig.\ref{step_bidir}. We can see from the graph that initially actuator takes two seconds to actuate completely, which is very fast and impressive. After holding the actuated state for about a second, the actuator releases the pressure to return to its starting state in two more seconds. Hence releasing is faster than actuation in this case. Overall, in little more than five seconds, the actuator completes the entire actuation cycle, making it suitable for most robotic applications.

\subsubsection{Pressure vs Displacement}

\begin{figure}[t]
    \centering
    \includegraphics[width=\linewidth]{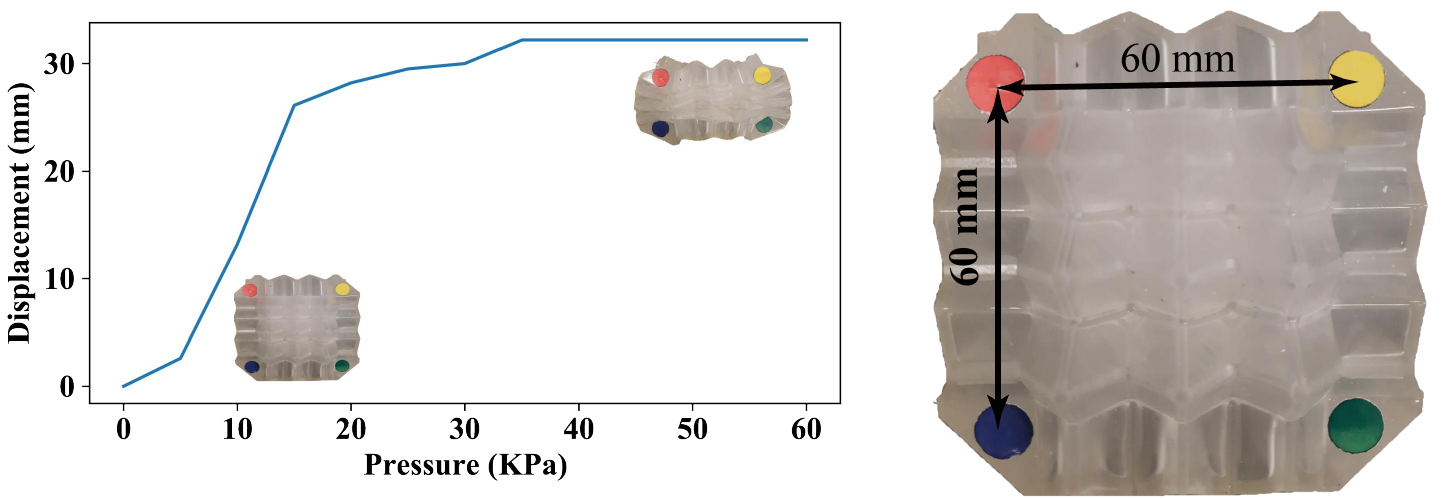}
    \caption{Relationship between applied vacuum and displacement of the actuator. Second image shows color dots marked on the surface for video analysis along with the dimension}
    \label{press_bidir}
\end{figure}

To assess the effect of vacuum pressure on the displacement of the actuator, we tested actuators with varying pressure inputs. Similar to the previous experiments, actuators are printed using the LCD 3D printing technique and Resinone F39T material. Material information is essential because inherent stiffness plays a crucial role in the movement of the actuator. The data is recorded visually in video format, which was then analyzed using Kinovea software; for better calibration, color dots were stuck on the surface (Fig.\ref{press_bidir}). The relationship between applied pressure and displacement is shown in Fig.\ref{press_bidir}, it can be concluded that the actuator reaches maximum change in length at around 35KPa negative pressure. 

\subsubsection{Shape of Trajectory and Directional Dependency}

\begin{figure}[t]
    \centering
    \includegraphics[width=\linewidth]{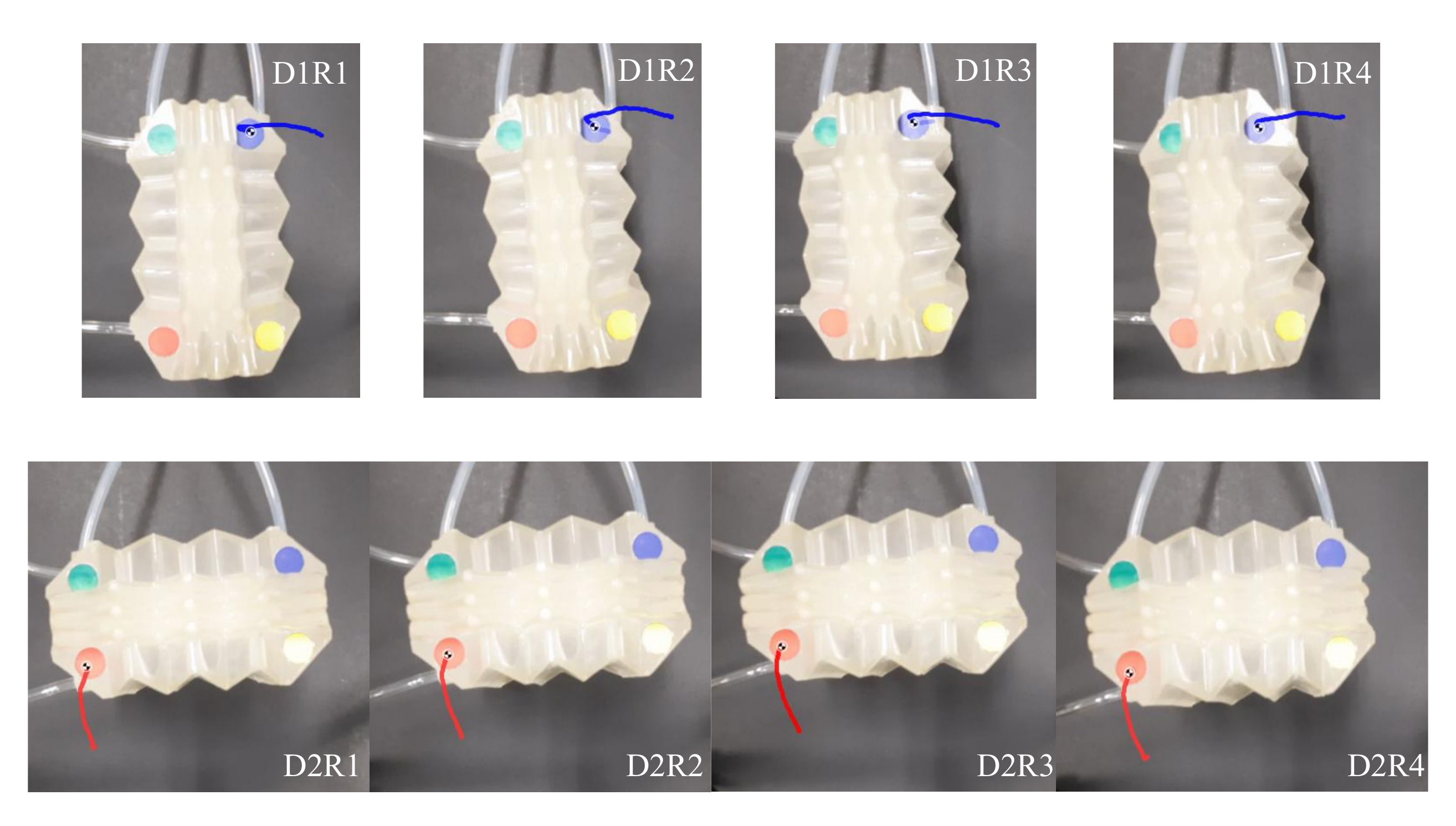}
    \caption{Experiment to analyse directional dependency of the actuator. The figures show actuator trajectories in two directions for four successive rounds (indicated in blue and red). D indicates the direction of the actuation and R is the round}
    \label{dependency}
\end{figure}

Since the fabrication material is elastic and soft, the intended orthogonal direction movements are not entirely straight; instead, they are curved. The curved path in both directions can be seen in the Fig.\ref{dependency}. However, despite this curvy-ness, the movement can be approximated as straight, and in a figurative sense, the motion is still orthogonal. 

In addition, as the actuation in one direction also causes movement in the second direction tubes, it is essential to analyze the effect of successive actuation. To quantify this, we actuated the bi-directional actuator in both directions alternatively for four rounds. The actuator trajectory for each round in both directions is shown in Fig.\ref{dependency}. It can be easily comprehended from these trajectories that for individual motion, there is little to no dependency between the two moving directions. We can easily compare and conclude that even if one direction's tubes are actuated right after the actuation in the other direction, the movement between the two will have a meager dependency. This can be explained by the fact that once the actuation in particular directions is released, all the cells in the design come to their initial state, and no deformation is left. Due to this, the second direction motion, even if it starts after the first direction motion, will start from a point where the actuator is in its normal state and, therefore, will not bear any correlation from the previous one. Nevertheless, it is equally important to mention that if somehow the material of the actuator degrades or fails, the in-dependency of the bi-directional motion will cease to exist.

\subsection{Modelling}
\subsubsection{Material Modelling}
\begin{figure}[t]
\centering
\includegraphics[width=\columnwidth]{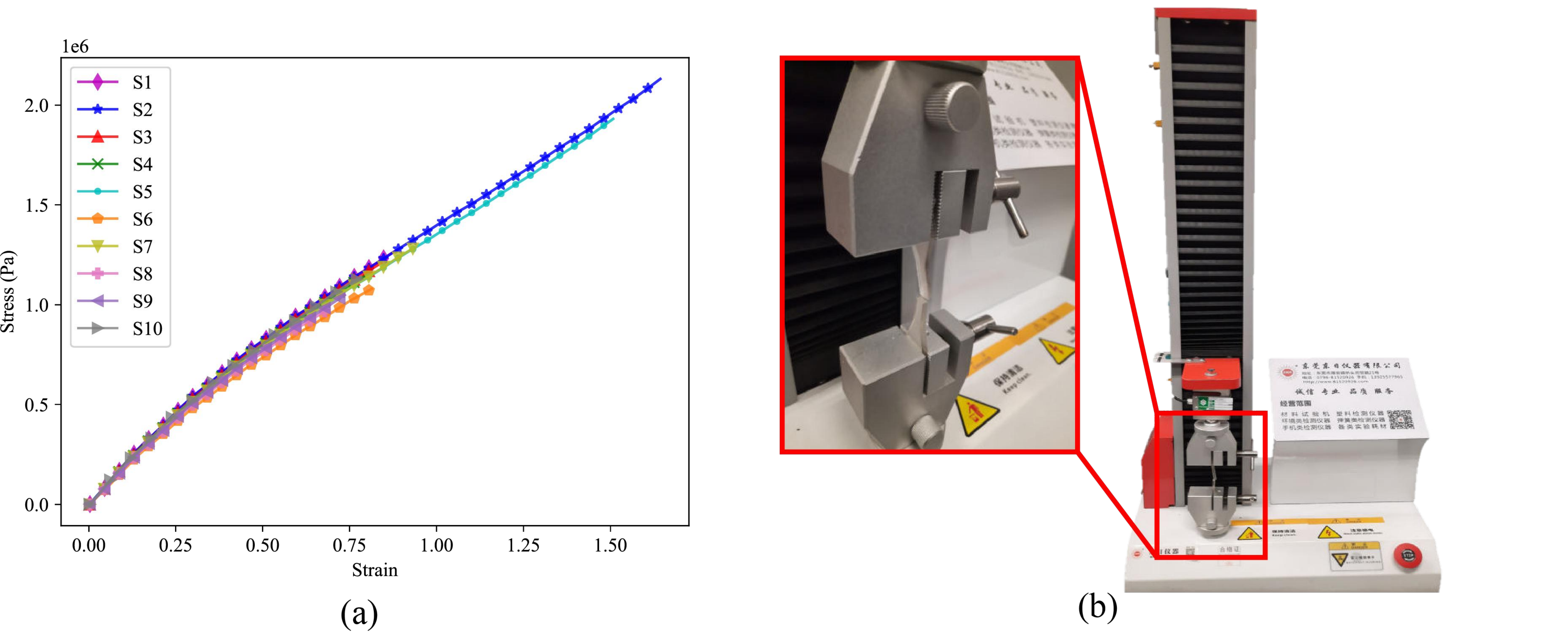}
\caption{(a) Stress-Strain relationship of the resin material, (b) UTM machine used for the material testing }
\label{UTMstress}
\end{figure}

A numerical or analytical model of the actuator will help us ascertain the forces generated and determine the expected displacements. Such information is very crucial to predict the behavior of the actuator and to test different design parameters. Unfortunately, material properties for elastic resin used to print our actuator were not widely available in the literature. The manufacturer did provide the datasheet, but that does not have many details. Therefore, we decided to derive the material's model in-house empirically using a Universal Testing Machine (UTM).  
UTM model DR502 from Dongri, with sensor type TFS-200N, is used for testing purposes (Fig.\ref{UTMstress}(b)). We did experimentation using ten specimens. All the specimens are printed using SLA resin on the form3 printer. Testing is done in compliance with the ASTM D638 Type IV standard. Specimens are printed in the dumbbell shape using the dimensions specified in \cite{astm}. The designated speed of testing for the mentioned Type is between 50-500 mm/min. The highest value is for the rigid material, and the lower values are for the softer materials. Our material was soft and elastic, and therefore, we used values between 50-70 mm/min for the experiments. Testing results in a nonlinear stress-strain relationship (Fig.\ref{UTMstress}), which the Ogden N=1 material model successfully fits. The resulting parameters for the Ogden model are $\mu_1$ = 708211.0002, $\alpha_1$ = 2.33765815 and $\textit{d}_1$ = 0.

\begin{figure}[t]
\centering
\includegraphics[width=\columnwidth]{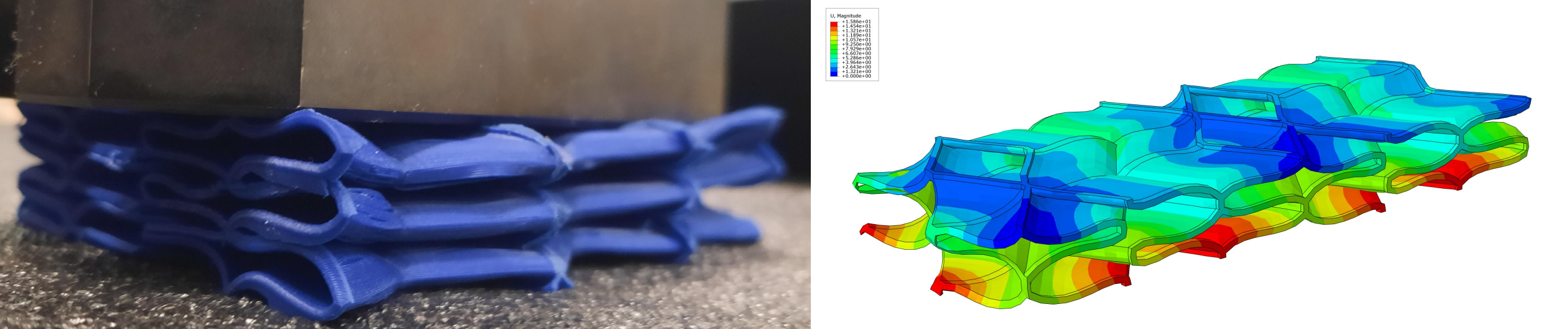}
\caption{Comparison between deformation of real and simulated model}
\label{abaqus22}
\end{figure}

\begin{figure}[t]
\centering
\includegraphics[width=\columnwidth]{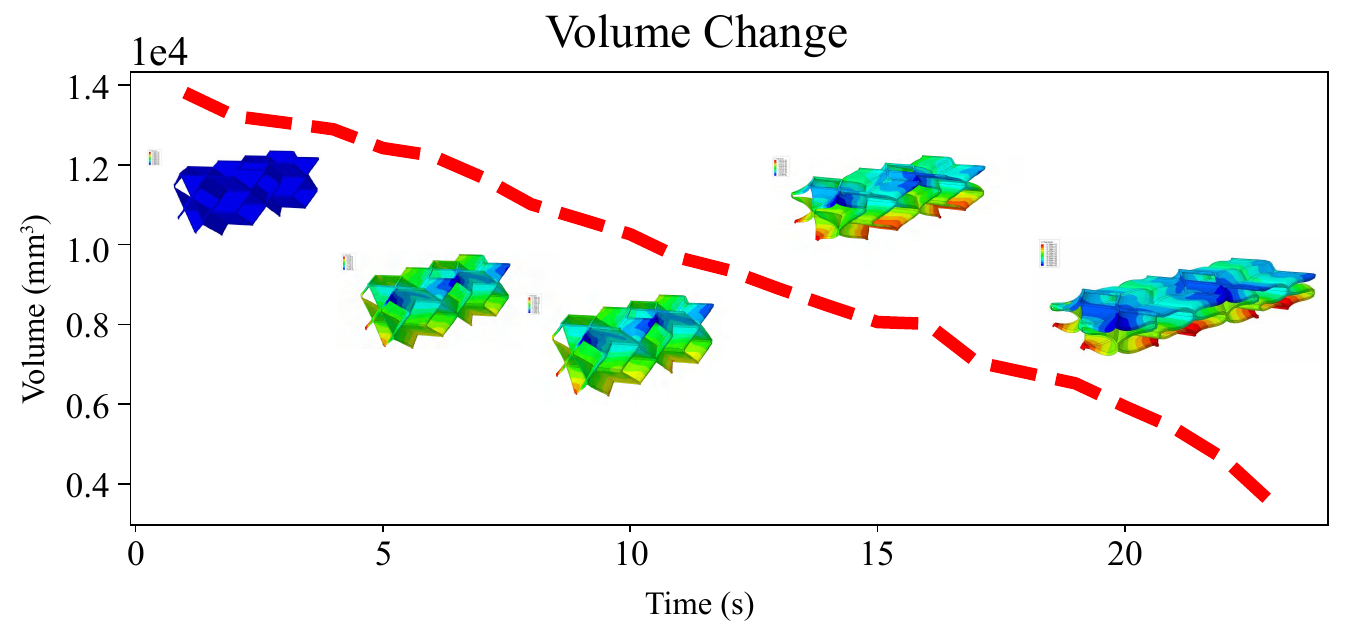}
\caption{Volume change of the inner tube structure during actuation}
\label{Volume Change}
\end{figure}

\subsubsection{FEM Modelling}

The finite element method (FEM) is a potent tool in determining actuators' simulated performance and predicting its behavior. FEM is widely used for the simulations of soft and semi-soft systems. Our study has formulated a nonlinear FEM model of the actuator to predict the folding behavior. We did simulations on a single tube and an entire 3D actuator to see how they fold when subjected to tensile loading. To make the simulation less expensive, we have simplified the design and only incorporated the core 3D structure and omitted the bellows on the side. 

Simulation is done using  Abaqus Simulia in static general step with Ngleom ON as the material is hyper-elastic.  The material model derived in the previous section was used in the study. Tetrahedral elements with Hybrid Formulation are used for meshing the surface. Self Contacts are defined between the surfaces that might come in contact during the simulation.

The behavior is simulated by applying different displacement boundary conditions at the ends of the actuators. We did not simulate the vacuum actuation due to the complex and large geometry inside the structure. Doing such a simulation would be computationally expensive and will also be hugely time-consuming. In addition, as the primary actuation in the structure is achieved through bending of the core structure and vacuum also exerts a net compressing force; therefore, we just performed the simulation of the scenario when the core structure is subjected to tensile loading. Due to the symmetry, we only tested one-fourth of the inner structure in the simulation study. The result for the one-fourth portion is shown in Fig.\ref{abaqus22}. The simulation model wholly captures the motion of the internal tube structures. For comparison, we have also depicted the actual design deformation when subjected to tensile loading. The variation of the actuator's inner volume is illustrated in Fig.\ref{Volume Change}. 

\section{APPLICATION AND EXPERIMENTS}
\subsection{Crawling Robot}
\begin{figure}[h]
    \centering
    \includegraphics[width=\columnwidth]{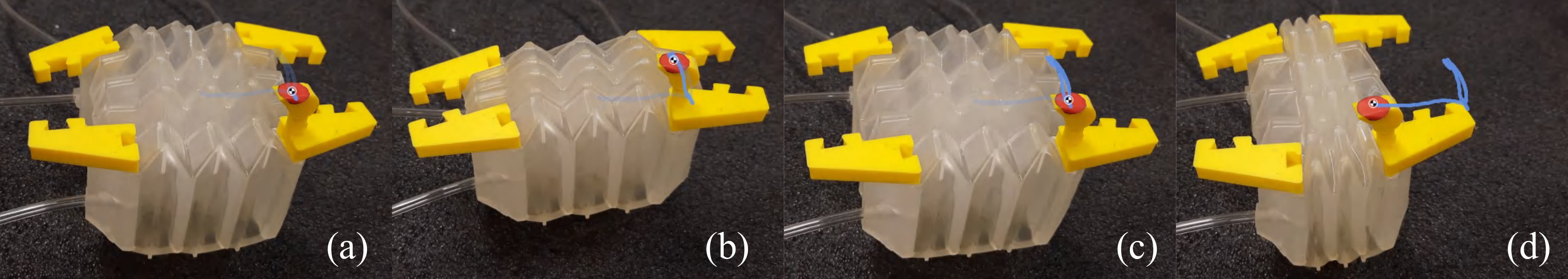}
    \caption{Demonstration of a orthogonal platform: (a) initial state, (b) movement in first direction, (c) back to initial state, (d) movement in the second direction (look at the "L" shaped trajectory)}
    \label{dualtest}
\end{figure}

To demonstrate the ability of the bi-directional actuator as a component for soft robotics here, we am presenting two concept designs. The first is of an orthogonal platform using which any object can be manipulated in orthogonal directions. Fig.\ref{dualtest} depicts the implementation for a small fingertip. We can see from the figure how the finger moves in an "L" shape trajectory. After scaling the actuator, even bigger parts can also be attached.

\begin{figure}[h]
    \centering
    \includegraphics[width=\columnwidth]{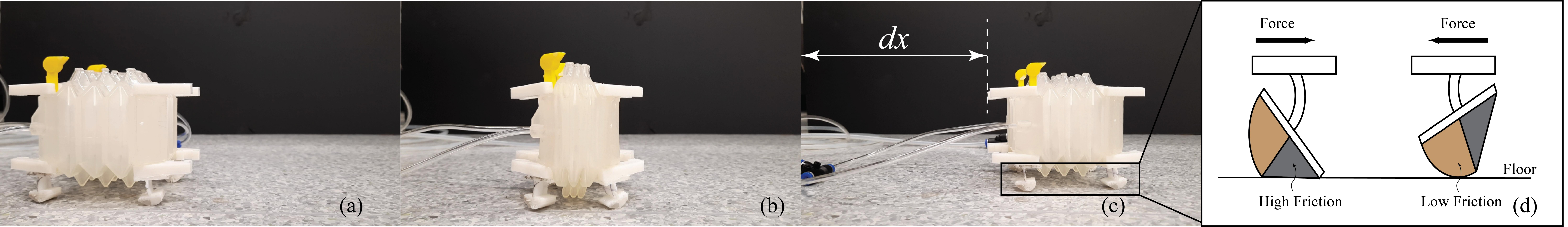}
    \caption{Demonstration of crawling robot; (a) initial state, (b) intermediate state, (c) final state, (d) explanation of anisotopic friction pads }
    \label{crawltest}
\end{figure}

The second concept design presented is of a crawling robot. Snapshots of the actuator crawling on the floor are shown in Fig.\ref{crawltest}, (a) is the initial state, (b) is the intermediate state, and (c) final state after crawling. For crawling to happen in the forward direction, we designed variable friction pads illustrated in Fig.\ref{crawltest}(d). The applied tangential force controls friction on the pad. Pads are arranged on the robot in such a way that the robot can crawl forward. PLA plastic part is used as low friction material, and silicon is used as a higher friction material. For the motion to happen in a particular direction, low friction tips should contact the ground on the front legs and high friction pads on the rear ones.
\begin{figure}[h]
    \centering
    \includegraphics[width=\columnwidth]{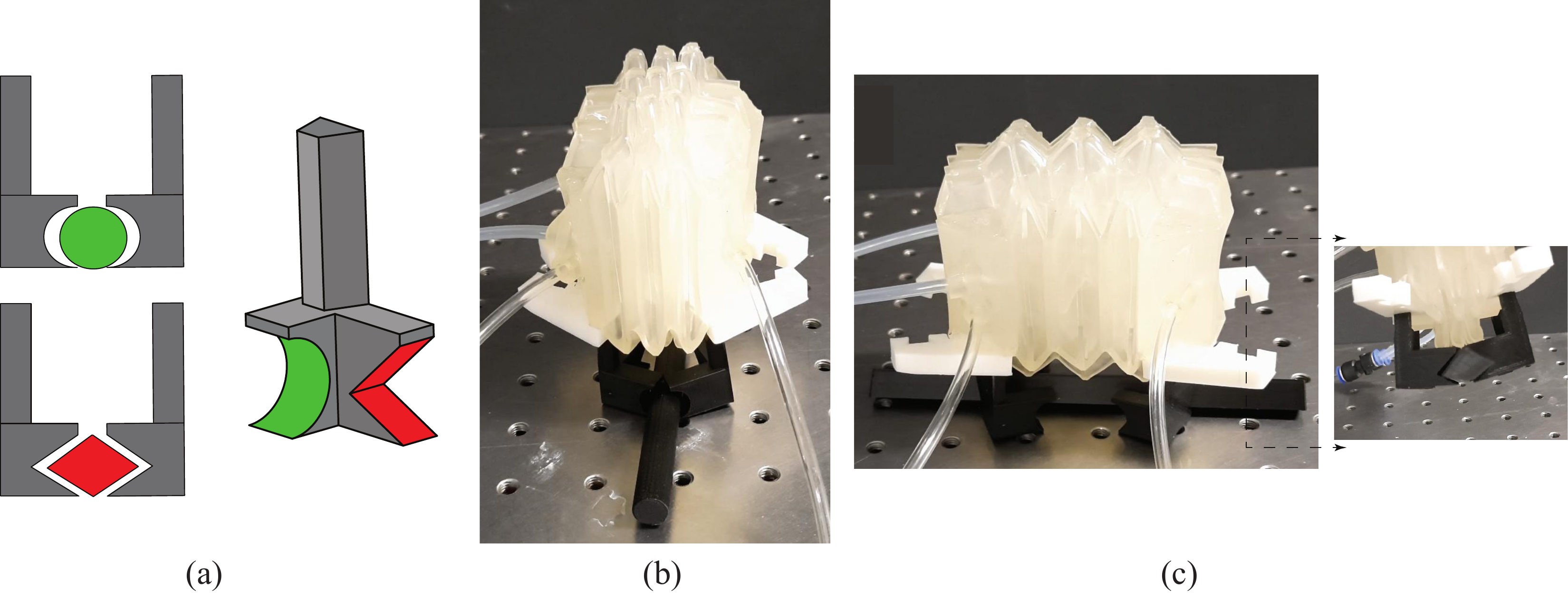}
    \caption{(a) Design of the gripper end effector/foot showing two different cross-sections that can be gripped, (b) Actuator gripping the round shaped objects, (c) Actuator gripping the diamond shaped objects. The last photo shows side view for better visualization }
    \label{dual_grip}
\end{figure}

\subsection{Bidirectional Platform}
\begin{figure}[h]
    \centering
    \includegraphics[width=0.75\columnwidth]{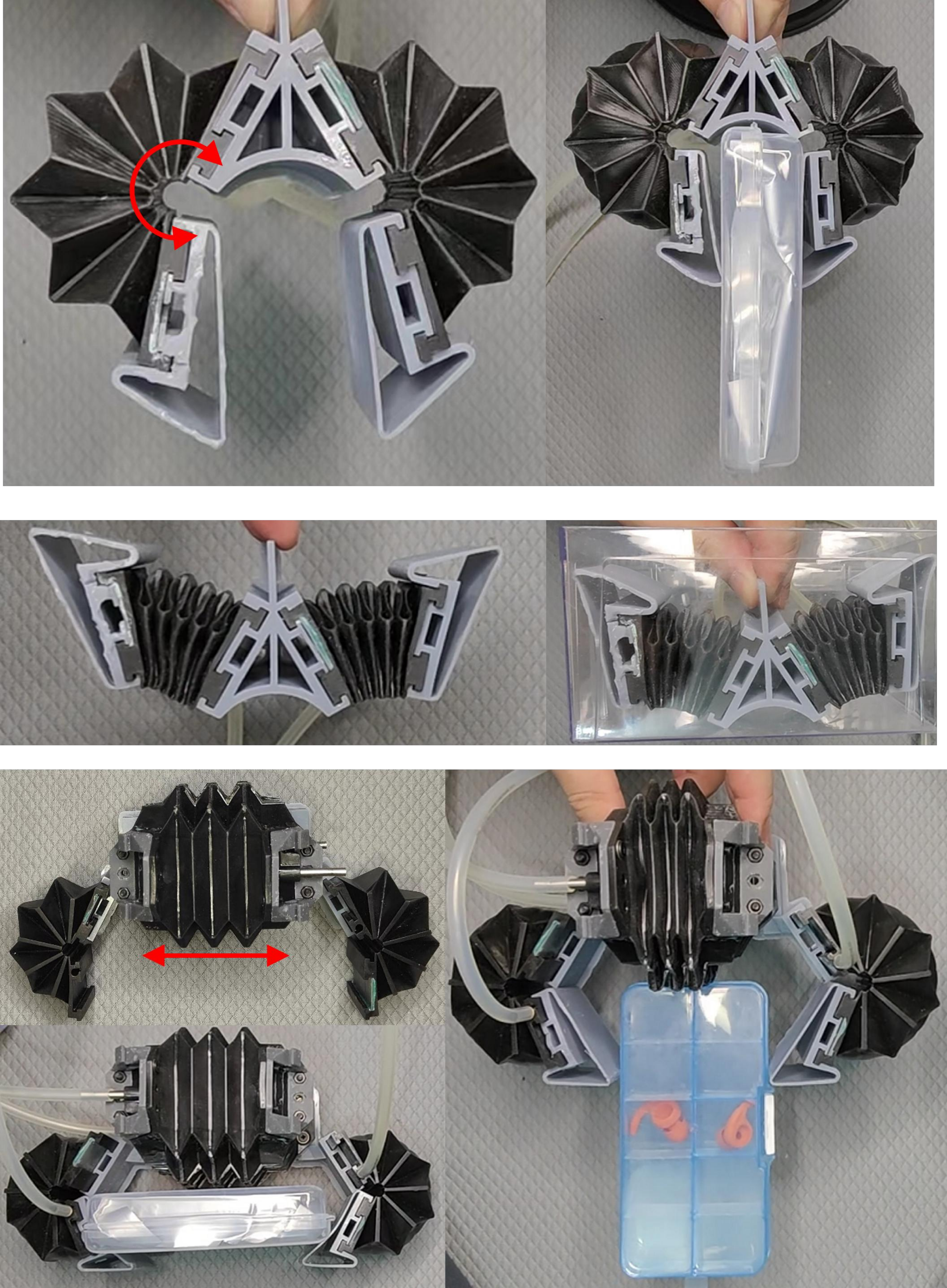}
    \caption{Combination of bi-directional actuators and rotary actuators to achieve different grasping capability.}
    \label{bi2}
\end{figure}

The bi-directional actuator provides two degrees of freedom for actuation, we can combine it with our proposed rotary actuator to achieve hybrid motion. As shown in Fig. \ref{bi2}, the rotary actuator itself can achieve dual-mode grasping: either from the outside of the object or from the inside of the object. When combined with the bi-directional actuator, it can adjust the gap distance between the two rotary actuators. In this way, the hybrid actuator can handle more objects in different scales.

\section{DISCUSSION}
This paper presents the design and development of a novel bi-directional origami actuator. This innovative structure integrates two vertically aligned driving chambers, achieving dual motions within a constrained volume. Our research journey involved extensive tests on various materials and fabrication techniques to enhance the success rate and durability of this origami structure. The culmination of our efforts resulted in a systematic approach to the design and production of this actuator. This design breakthrough opens a new horizon for researchers, offering fresh perspectives and inspirations in crafting unique actuators using origami structures. Our work not only contributes to the field theoretically but also has practical implications. We envision this bi-directional origami actuator to be utilized in applications like quadruped robot propulsion and mechanical gripper actuation. The adaptability of our design can cater to a wide range of robotic applications, underscoring its potential in advancing the domain of soft robotics. 
However, despite the promising results from our tests, the actuator currently exhibits several challenges:
1. Under heavy loads, the direction of the actuator's movement tends to deviate from the desired trajectory. This is primarily due to the initial air pressure being insufficient to generate effective startup displacement.
2. The control of the actuator remains a significant challenge. Given the material non-linearity and the complexity of the structure, currently, it can only operate in binary modes, i.e., either open or shut.
Addressing these issues will be pivotal to ensure the seamless integration and application of this actuator in real-world scenarios. The journey of perfecting this novel origami-inspired actuator is ongoing, and the lessons learned from the current challenges will undoubtedly fuel future refinements and innovations in the field.

\section{ACKNOWLEDGEMENT}

This research work is supported by the Hong Kong General Research Fund, with the project No. RGC16200919.

\bibliographystyle{IEEEtran}
\bibliography{reference}

\begin{thebibliography}{10}
\providecommand{\url}[1]{#1}
\csname url@samestyle\endcsname
\providecommand{\newblock}{\relax}
\providecommand{\bibinfo}[2]{#2}
\providecommand{\BIBentrySTDinterwordspacing}{\spaceskip=0pt\relax}
\providecommand{\BIBentryALTinterwordstretchfactor}{4}
\providecommand{\BIBentryALTinterwordspacing}{\spaceskip=\fontdimen2\font plus
\BIBentryALTinterwordstretchfactor\fontdimen3\font minus \fontdimen4\font\relax}
\providecommand{\BIBforeignlanguage}[2]{{%
\expandafter\ifx\csname l@#1\endcsname\relax
\typeout{** WARNING: IEEEtran.bst: No hyphenation pattern has been}%
\typeout{** loaded for the language `#1'. Using the pattern for}%
\typeout{** the default language instead.}%
\else
\language=\csname l@#1\endcsname
\fi
#2}}
\providecommand{\BIBdecl}{\relax}
\BIBdecl

\bibitem{rus2018design}
D.~Rus and M.~T. Tolley, ``Design, fabrication and control of origami robots,'' \emph{Nature Reviews Materials}, vol.~3, no.~6, pp. 101--112, 2018.

\bibitem{boyvat2017addressable}
M.~Boyvat, J.-S. Koh, and R.~J. Wood, ``Addressable wireless actuation for multijoint folding robots and devices,'' \emph{Science Robotics}, vol.~2, no.~8, 2017.

\bibitem{wilson2013origami}
L.~Wilson, S.~Pellegrino, and R.~Danner, ``Origami sunshield concepts for space telescopes,'' in \emph{54th AIAA/ASME/ASCE/AHS/ASC Structures, Structural Dynamics, and Materials Conference}, 2013, p. 1594.

\bibitem{yao2019reconfiguration}
M.~Yao, C.~H. Belke, H.~Cui, and J.~Paik, ``A reconfiguration strategy for modular robots using origami folding,'' \emph{The International Journal of Robotics Research}, vol.~38, no.~1, pp. 73--89, 2019.

\bibitem{peraza2014origami}
E.~A. Peraza-Hernandez, D.~J. Hartl, R.~J. Malak~Jr, and D.~C. Lagoudas, ``Origami-inspired active structures: a synthesis and review,'' \emph{Smart Materials and Structures}, vol.~23, no.~9, p. 094001, 2014.

\bibitem{defigueiredo2019origami}
B.~P. DeFigueiredo, N.~A. Pehrson, K.~A. Tolman, E.~Crampton, S.~P. Magleby, and L.~L. Howell, ``Origami-based design of conceal-and-reveal systems,'' \emph{Journal of Mechanisms and Robotics}, vol.~11, no.~2, 2019.

\bibitem{lee2013deformable}
D.-Y. Lee, G.-P. Jung, M.-K. Sin, S.-H. Ahn, and K.-J. Cho, ``Deformable wheel robot based on origami structure,'' in \emph{2013 IEEE International Conference on Robotics and Automation}.\hskip 1em plus 0.5em minus 0.4em\relax IEEE, 2013, pp. 5612--5617.

\bibitem{koh2012omega}
J.-S. Koh and K.-J. Cho, ``Omega-shaped inchworm-inspired crawling robot with large-index-and-pitch (lip) sma spring actuators,'' \emph{IEEE/ASME Transactions On Mechatronics}, vol.~18, no.~2, pp. 419--429, 2012.

\bibitem{sreetharan2012monolithic}
P.~S. Sreetharan, J.~P. Whitney, M.~D. Strauss, and R.~J. Wood, ``Monolithic fabrication of millimeter-scale machines,'' \emph{Journal of Micromechanics and Microengineering}, vol.~22, no.~5, p. 055027, 2012.

\bibitem{firouzeh2017under}
A.~Firouzeh and J.~Paik, ``An under-actuated origami gripper with adjustable stiffness joints for multiple grasp modes,'' \emph{Smart Materials and Structures}, vol.~26, no.~5, p. 055035, 2017.

\bibitem{weston2017towards}
W.~P. Weston-Dawkes, A.~C. Ong, M.~R.~A. Majit, F.~Joseph, and M.~T. Tolley, ``Towards rapid mechanical customization of cm-scale self-folding agents,'' in \emph{2017 IEEE/RSJ International Conference on Intelligent Robots and Systems (IROS)}.\hskip 1em plus 0.5em minus 0.4em\relax IEEE, 2017, pp. 4312--4318.

\bibitem{edmondson2013oriceps}
B.~J. Edmondson, L.~A. Bowen, C.~L. Grames, S.~P. Magleby, L.~L. Howell, and T.~C. Bateman, ``Oriceps: Origami-inspired forceps,'' in \emph{ASME 2013 conference on smart materials, adaptive structures and intelligent systems}.\hskip 1em plus 0.5em minus 0.4em\relax American Society of Mechanical Engineers Digital Collection, 2013.

\bibitem{santoso2020origami}
J.~Santoso and C.~D. Onal, ``An origami continuum robot capable of precise motion through torsionally stiff body and smooth inverse kinematics,'' \emph{Soft Robotics}, 2020.

\bibitem{10.1115/DETC2013-13490}
\BIBentryALTinterwordspacing
\emph{{Conceptual Model Study Using Origami for Membrane Space Structures}}, ser. International Design Engineering Technical Conferences and Computers and Information in Engineering Conference, vol. Volume 6B: 37th Mechanisms and Robotics Conference, 08 2013, v06BT07A047. [Online]. Available: \url{https://doi.org/10.1115/DETC2013-13490}
\BIBentrySTDinterwordspacing

\bibitem{kim2018origami}
S.-J. Kim, D.-Y. Lee, G.-P. Jung, and K.-J. Cho, ``An origami-inspired, self-locking robotic arm that can be folded flat,'' \emph{Science Robotics}, vol.~3, no.~16, 2018.

\bibitem{li2017fluid}
S.~Li, D.~M. Vogt, D.~Rus, and R.~J. Wood, ``Fluid-driven origami-inspired artificial muscles,'' \emph{Proceedings of the National academy of Sciences}, vol. 114, no.~50, pp. 13\,132--13\,137, 2017.

\bibitem{filipov2015origami}
E.~T. Filipov, T.~Tachi, and G.~H. Paulino, ``Origami tubes assembled into stiff, yet reconfigurable structures and metamaterials,'' \emph{Proceedings of the National Academy of Sciences}, vol. 112, no.~40, pp. 12\,321--12\,326, 2015.

\bibitem{filipov2016origami}
E.~T. Filipov, G.~Paulino, and T.~Tachi, ``Origami tubes with reconfigurable polygonal cross-sections,'' \emph{Proceedings of the Royal Society A: Mathematical, Physical and Engineering Sciences}, vol. 472, no. 2185, p. 20150607, 2016.

\bibitem{chen2017extended}
Y.~Chen, W.~Lv, J.~Li, and Z.~You, ``An extended family of rigidly foldable origami tubes,'' \emph{Journal of Mechanisms and Robotics}, vol.~9, no.~2, 2017.

\bibitem{astm}
``D638-14 standard test method for tensile properties of plastics,'' \emph{ASTM International}, 2014.

\end{thebibliography}

\end{document}